\documentclass[10pt, a4paper]{article}
\usepackage{lrec2022} 
\usepackage{multibib}
\newcites{languageresource}{Language Resources}
\usepackage{graphicx}
\usepackage{tabularx}
\usepackage{soul}
\usepackage{titlesec}
\titleformat{\section}{\normalfont\large\bfseries\center}{\thesection.}{1em}{}
\titleformat{\subsection}{\normalfont\SmallTitleFont\bfseries\raggedright}{\thesubsection.}{1em}{}
\titleformat{\subsubsection}{\normalfont\normalsize\bfseries\raggedright}{\thesubsubsection.}{1em}{}
\renewcommand\thesection{\arabic{section}}
\renewcommand\thesubsection{\thesection.\arabic{subsection}}
\renewcommand\thesubsubsection{\thesubsection.\arabic{subsubsection}}

\usepackage{epstopdf}
\usepackage[utf8]{inputenc}

\usepackage{hyperref}
\usepackage{xstring}

\usepackage{color}

\usepackage[utf8]{inputenc} 
\usepackage[T1]{fontenc}    
\usepackage{hyperref}       
\usepackage{url}            
\usepackage{booktabs}       
\usepackage{amsfonts}       
\usepackage{nicefrac}       
\usepackage{microtype}      
\usepackage{xcolor}         
\usepackage{array}          
\usepackage{longtable}      
\usepackage{makecell}       
\usepackage{graphicx}       
\usepackage{subcaption}     
\usepackage{authblk}
\usepackage{times}
\usepackage{latexsym}
\usepackage[utf8]{inputenc} 
\usepackage[T1]{fontenc}    
\usepackage{hyperref}       
\usepackage{url}            
\usepackage{booktabs}       
\usepackage{amsfonts}       
\usepackage{nicefrac}       
\usepackage{microtype}      
\usepackage{xcolor}         
\usepackage{array}          
\usepackage{longtable}      
\usepackage{makecell}       
\usepackage{graphicx}       
\usepackage{subcaption}     
\newcolumntype{M}[1]{>{\centering\arraybackslash}m{#1}} 

\title{DrugEHRQA: A Question Answering Dataset on Structured and Unstructured Electronic Health Records For Medicine Related Queries }

\name{Jayetri Bardhan$^{1\dagger}$, Anthony Colas$^{1\ddagger}$, Kirk Roberts$^{2\ast}$, Daisy Zhe Wang$^{1\diamond}$} 

\address{$^1$Department of Computer and Information Science and Engineering, 
  University of Florida \\
         $^2$School of Biomedical Informatics, The University of Texas Health Science Center at Houston \\
         \{jayetri.bardhan$^{\dagger}$, acolas1$^{\ddagger}$, daisyw$^{\diamond}$\}@ufl.edu, kirk.roberts@uth.tmc.edu$^{\ast}$}

\abstract{
This paper develops the first question answering dataset (DrugEHRQA) containing question-answer pairs from both structured tables and unstructured notes from a publicly available Electronic Health Record (EHR). EHRs contain patient records, stored in structured tables and unstructured clinical notes. The information in structured and unstructured EHRs is not strictly disjoint: information may be duplicated, contradictory, or provide additional context between these sources. Our dataset has medication-related queries, containing over 70,000 question-answer pairs. To provide a baseline model and help analyze the dataset, we have used a simple model (MultimodalEHRQA) which uses the predictions of a modality selection network to choose between EHR tables and clinical notes to answer the questions. This is used to direct the questions to the table-based or text-based state-of-the-art QA model. In order to address the problem arising from complex, nested queries, this is the first time Relation-Aware Schema Encoding and Linking for Text-to-SQL Parsers (RAT-SQL) has been used to test the structure of query templates in EHR data. Our goal is to provide a benchmark dataset for multi-modal QA systems, and to open up new avenues of research in improving question answering over EHR structured data by using context from unstructured clinical data.
 \\ \newline  }

\begin{document}

\maketitleabstract

\section{Introduction}

Electronic Health Records (EHRs) are digitized records of patients’ medical history, which can be either in structured or unstructured form. MIMIC-III (Medical Information Mart for Intensive Care)~\cite{1.4johsonmimic} is an example of a large EHR database containing health-related information for over 40,000 patients who were admitted in critical care units in the period between 2001 and 2012 in the Beth Israel Deaconess Medical Center. MIMIC-III contains multi-relational tables with patient-specific data on diagnosis, medications, admission, lab tests, etc. Question answering over EHRs aid doctors in diagnosing, while helping patients obtain answers to health-related queries. The structured relational database has multiple tables which store information about the patient’s demographics, diagnoses, medications, lab tests along with their results. Conversely, the unstructured data, are notes entered by clinicians that contain a  detailed  description of every patient’s visit, their past medical history, their problem, symptoms and more. Thus, to benefit from both the modalities, there arises a  need for a multi-modal QA dataset on EHRs.

We present DrugEHRQA, the first QA dataset which uses both the structured tables and the unstructured clinical notes of an EHR to answer questions. The answers from the clinical notes are used to support or provide evidence to the answers retrieved from the structured tables. The former gives better context to support the latter. Moreover, there can be cases where a guaranteed answer might not be available in the structured tables, due to missing data/relation. For example, if the question is ‘What medication is the patient with an admission ID of 105104 taking for Hypoxemia?’ The MIMIC-III tables have no direct relation between medicines and problems. The tables DIAGNOSES\_ICD and D\_ICD\_DIAGNOSES of MIMIC-III can be used to verify if the patient with admission ID 105104 is suffering from Hypoxemia, and the PRESCRIPTIONS table of MIMIC-III can be used to fetch all the medicines prescribed to the patient, having an admission ID of 105104. However, the patient could have been prescribed medicines for non-Hypoxemia related conditions, which will be contained in the tables. In such cases, the unstructured clinical notes can be used to identify the medicines from this list, since the information about the medicine for Hypoxemia is directly present in the clinical notes.

One reason for the lack of any pre-existing multimodal QA dataset on EHRs is due to the tedious amount of time and effort that is required to annotate such a dataset. In this work, we introduce a novel method to automatically generate a template-based QA (DrugEHRQA) dataset (on drug-related questions) from the MIMIC-III database. DrugEHRQA contains the following: 1) natural language questions, 2) its corresponding SQL Query that can be used to retrieve answers from the multi-relational MIMIC-III tables, 3) the answers from either or both the modalities, and 4) the ‘best selected’ multi-modal answer. DrugEHRQA contains 70,381 QA pairs that have been generated using nine different template types. We also generated three paraphrases for every natural language question template, and analyzed the effects of paraphrasing on the baseline models.  DrugEHRQA was benchmarked against existing models like  TREQS~\cite{wang2020text}, RAT-SQL~\cite{wang2020rat}, BERT QA~\cite{devlin2019bert} and ClinicalBERT~\cite{alsentzer2019publicly} to test the validity of the DrugEHRQA dataset for the individual modalities.

We also introduce a simple baseline model for multimodal text-table QA in EHRs. Similar to ManyModelQA~\cite{hannan2020manymodalqa}, our model (MultimodalEHRQA) contains a modality selection network. With questions fed as an input, the modality selection network predicts the more reliable modality (i.e. text or table) for question answering. The predicted modality is used to choose the corresponding state-of-the-art model for table and text QA. The more complex task of leveraging answers from both structured and unstructured data to provide a contextualized answer is left to future work.  

The main contributions of this paper are as follows:
\begin{enumerate}
    \item Introduce DrugEHRQA\footnote{https://github.com/jayetri/DrugEHRQA-A-Question-Answering-Dataset-on-Structured-and-Unstructured-Electronic-Health-Records, scripts to generate DrugEHRQA dataset}, the first QA dataset on multi-modal EHRs, containing QA pairs from structured tables and unstructured clinical notes from MIMIC III. The dataset contains natural language questions, its corresponding SQL query for querying multi-relational tables in MIMIC-III, the retrieved answer(s) from one or both modalities, and the combined multi-modal answer.  
     
    \item Introduce a novel technique to automatically generate a template-based dataset using existing annotations of a non-QA application, skipping the need for manual annotations specifically for QA.
    \item Develop a simple baseline model for multimodal QA on EHR, using modality selection network.
    
\end{enumerate}

The remainder of the paper is organized into 8 sections. Section~\ref{related_work} discusses existing related work, Section~\ref{dataset_gen} describes the DrugEHRQA dataset generation, Section~\ref{analysis_DrugEHRQA} presents the analysis of DrugEHRQA, Section~\ref{models} discusses the baseline model for multimodal QA, Section~\ref{experiments_and_results} presents the experiments and results, Section~\ref{limitations} discusses the reproducibility and limitations of our work, Section~\ref{broader_impact} proposes the broader impact of our dataset in the EHR QA research community, and Section~\ref{conclusion and Future Work} concludes the work and discusses possible future work.

\section{Related Work}
\label{related_work}
Even though there are some available multimodal QA datasets in non-clinical domains~\cite{hannan2020manymodalqa,chen2020hybridqa,talmor2021multimodalqa}, but there are no existing multimodal QA datasets which uses structured with unstructured EHR data to answer questions. There are some existing works in the clinical genre on multi-modal understanding from text-image pairs ~\cite{moon2021multi,khare2021mmbert,li2020comparison} as well as clinical QA ~\cite{singh2021mimoqa} on text-image data. But as far as the authors' knowledge, so far there is no multi-modal clinical dataset that encorporates structured and unstructured EHR data for QA. QA in EHRs has been limited to QA over knowledge bases~\cite{wang2021attention}, EHR tables~\cite{wang2020text,raghavan2021emrkbqa} or clinical notes~\cite{johnson2016mimic,pampari2018emrqa}. emrQA~\cite{pampari2018emrqa} and CliniQG4QA~\cite{yue2021cliniqg4qa} are QA datasets that utilize unstructured text of EHRs to generate QA datasets. The emrQA contains 1 million question-logical forms along with over 40,000 QA evidence pairs, extracted from clinical notes of five n2c2 challenge  datasets \footnote{https://www.i2b2.org/NLP/DataSets/}. CliniQG4QA on the other hand, contains 1287 annotated QA pairs on 36 discharge summaries from clinical notes of MIMIC-III. CliCR~\cite{vsuster2018clicr} is another large medical QA dataset which is constructed from clinical case reports. It is used for reading comprehension in the healthcare domain. The reports used in CliCR are proxy for electronic health records, since the clinical reports look similar to the discharge summaries of EHR even though it lacks features of EHR text.

There are QA datasets that are generated using template-based method like MIMICSQL~\cite{wang2020text} and emrKBQA~\cite{raghavan2021emrkbqa} which utilize the structured EHR tables of MIMIC-III for QA. emrKBQA contains 940,000 questions, logical forms and answers which uses the structured records of MIMIC-III. Both emrKBQA and emrQA use semi-automated methods to retrieve the answers. The question templates and logical forms are generated by physicians, followed by a slot-filling process and answers retrieved from MIMIC-III KB~\cite{johnson2016mimic}.
On the contrary, our dataset - DrugEHRQA uses both structured tables and clinical notes containing elaborate details of MIMIC-III to generate the QA dataset. We use an automatic novel methodology to create the dataset (described in Section~\ref{dataset_gen}). 

\section{Dataset Generation}
\label{dataset_gen}
The dataset has been generated using a template-based method. The dataset (DrugEHRQA) contains over 70,000 natural language questions. Each line in DrugEHRQA consists of a natural language question, its corresponding SQL query to retrieve answers from the MIMIC-III tables, the retrieved answers from MIMIC-III tables and/or answers from clinical notes of MIMIC-III, and the selected multi-modal answer. As stated earlier, generating a multi-modal dataset is time-consuming mainly because the data must be manually annotated, which is a very tedious process. To overcome this, we  introduce a novel strategy to automatically generate the dataset. The dataset generation framework of DrugEHRQA is illustrated in  figure~\ref{fig:dataset_generation_framework}. The dataset generation process can be explained using five steps: (1) Annotation of question templates, (2) Extraction of drug based relations from n2c2 repository, (3) Answer extraction from MIMIC-III tables, (4) Paraphrasing, (5) Selecting multi-modal answers. The following subsections explain in detail the five steps involved in automatic data generation.  

\subsection{Annotation of Question Templates}
\label{annot_ques_temp}
We have annotated nine natural-language (NL) medicine-related question templates along with their corresponding SQL query templates. Five out of the nine NL question templates are taken from the medicine related templates of emrQA~\cite{pampari2018emrqa}.  The question templates are designed in such a way that their information appears in both structured and unstructured MIMIC-III data. The questions in the templates cover topics such as \emph{drug-dosage, drug strength, route, form of medicine, problems}. Table~\ref{tab:app_tab1} in the Appendix section shows the nine templates that have been used in the process of data generation. Each SQL query template is categorized into various difficulty levels- “easy”, “medium”, “hard” and “very hard”. The difficulty level is assigned based on the complexity of the SQL query, which is determined by number of "WHERE" conditions, the number of aggregation columns, presence/absence of aggregation operators, "GROUP BY", "ORDER BY", "LIMIT", number of tables, "JOINS" and nesting. For example, the SQL query template in the first row of the table~\ref{tab:app_tab1} is “easy” since it just has one aggregation column and one "WHERE" condition. But the SQL query template in the last row is nested, contains "JOINS" and has multiple "WHERE" conditions. Hence, it is classified as “very hard”. In the following sections, we use the terms “drug problems” and “drug reasons” interchangeably. This is because the data in the dataset is annotated as “drug reasons”, but to provide contextual clarity we use “drug problems” in this paper. 

\subsection{Answer Retrieval from Unstructured Data}
\label{ans_retrieval}

\begin{figure*}[hbt!]
    \centering
    \includegraphics[scale=0.56]{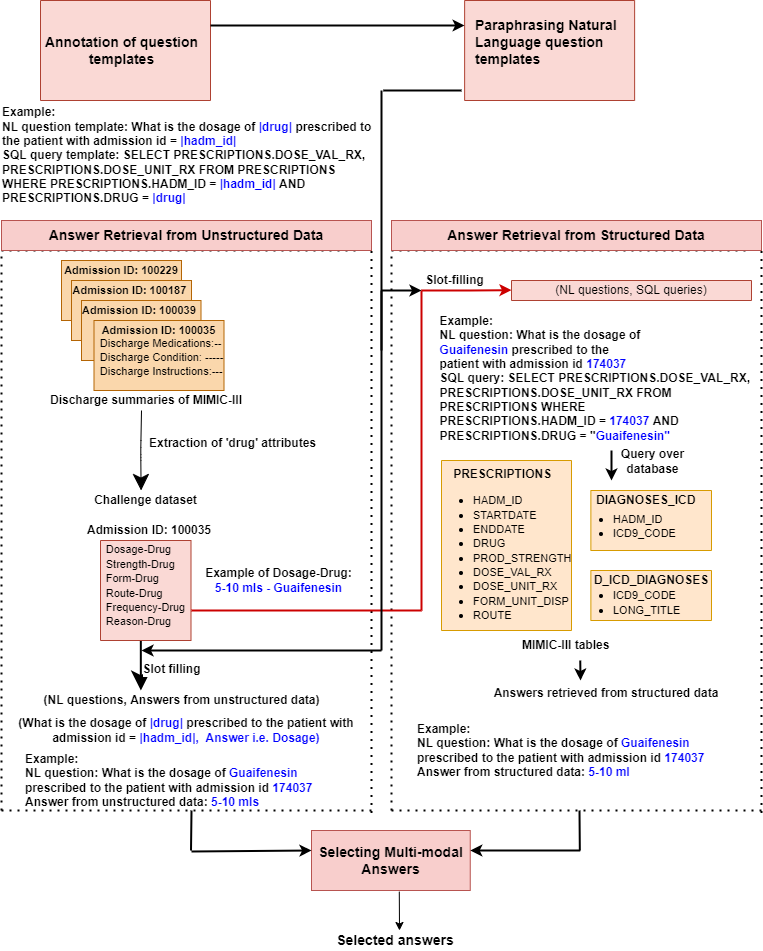}
    \caption{Dataset generation framework of DrugEHRQA. There are five steps in this process: (1) annotation of question templates, (2) answer retrieval from unstructured clinical notes, (3) answer retrieval from structured EHR Data, (4) paraphrasing natural language question templates, and (5) selecting multi-modal answers. Note that the challenge dataset mentioned in the figure refers to the "The 2018 Adverse Drug Event (ADE) dataset and Medication Extraction Challenge dataset" present the n2c2 repository.}
    \vspace{-1em}
    \label{fig:dataset_generation_framework}
\end{figure*}

"The 2018 Adverse Drug Event (ADE) dataset and Medical Extraction Challenge dataset"~\cite{henry20202018} present in the n2c2 repository\footnote{https://portal.dbmi.hms.harvard.edu/projects/n2c2-nlp/} contains annotations for 505 clinical notes of patients (from the MIMIC-III database), who had experienced ADE while they were admitted in the hospital. This dataset will be henceforth referred to as challenge dataset. We used the annotations from the challenge dataset to extract all the drug related attributes for the 505 discharge summaries of patients in the MIMIC-III database. We used six drug-related attributes, namely, Strength-Drug, Form-Drug, Route-Drug, Dosage-Drug, Frequency-Drug, and Reason-Drug, from the challenge dataset to generate QA pairs. We used each of these drug attributes and the medicine names to generate nine types of natural language question templates. For example, the annotation from Dosage-Drug for a certain admission ID is used to answer the question - "What is the dosage of |drug| prescribed to the patient with admission id = |hadm\_id|?", where |hadm\_id| refer to the admission ID of the patient. This is depicted in the figure~\ref{fig:dataset_generation_framework}. Table~\ref{tab:NL_question_template} lists the drug attributes with examples and its derived NL questions. The medicines, drug attributes and admission IDs of the 505 annotation files are slot-filled to replace the placeholders in the question templates to generate the question-answer pairs. For data licensing issues of n2c2 repository, we submitted this QA dataset on clinical notes of MIMIC-III to the n2c2 repository.

\subsection{Answer Extraction from MIMIC-III Tables}
\label{ans_extract_mimic3}
Extraction of answers from MIMIC-III tables is achieved by using the admission IDs, names of drugs and problems, utilized in the data generation process from unstructured data (Section~\ref{ans_retrieval}), to fill up the slots for |hadm\_id|, |drug| and |problem| in the NL and SQL Query templates (Section\ref{annot_ques_temp}). A slot filling process was used to generate the SQL queries that helped in retrieving answers from the MIMIC-III's structured database (refer to figure ~\ref{fig:dataset_generation_framework}). The answer may or may not exist in the MIMIC-III tables for the questions corresponding to the different combination of 505 admission IDs and entities of drugs (or problems) obtained from the clinical notes, resulting in an empty answer for certain questions. Three MIMIC-III tables, namely, PRESCRIPTIONS, DIAGNOSES\_ICD, and D\_ICD\_DIAGNOSES are used for data retrieval. The PRESCRIPTIONS table of MIMIC-III contains drug-related information, whereas the tables - DIAGNOSES\_ICD and D\_ICD\_DIAGNOSES contain the diagnosed results of the patients. The DrugEHRQA dataset now contains NL Questions, its corresponding SQL queries for querying the structured database, the answers retrieved from the structured tables (Answer Structured), and the answers retrieved from unstructured data (Answer Unstructured). 

\subsection{Paraphrasing Natural Language Questions}
\label{paraphrase_nat_lang_quest}
\begin{figure*}[ht]
    \centering
    \includegraphics[scale=0.3]{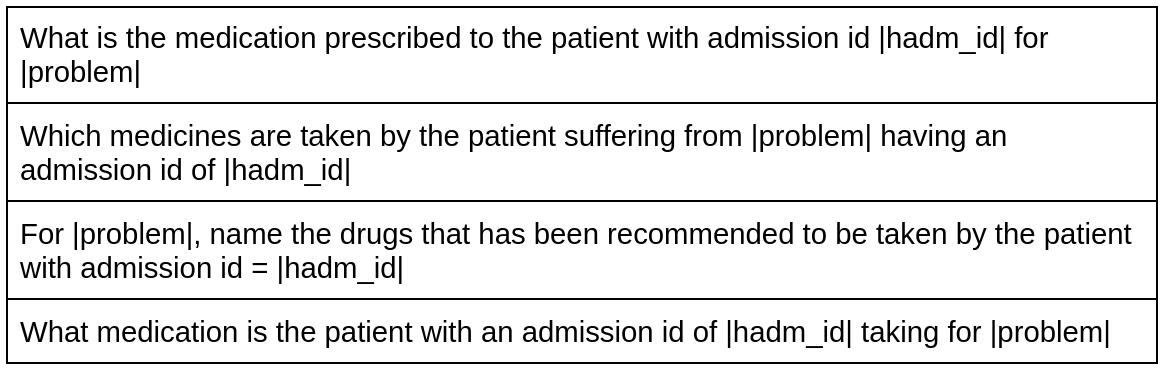}
    \caption{Example of various paraphrases of a natural language question template in the DrugEHRQA dataset.}
    \vspace{-1em}
    \label{fig:paraphrase_example}
\end{figure*}

Patients and clinicians may pose the same question in different formats (paraphrases). There has been a substantial amount of work done in EHR QA, studying the effects of NL paraphrasing in QA ~\cite{wang2020text,pampari2018emrqa,rawat2020entity,soni2019paraphrase,moon2020you}. We added paraphrases in the natural language question templates to improve the diversity of DrugEHRQA, making it more realistic and robust.  We created four paraphrases for each of the nine natural language query templates (i.e. three additional paraphrases per template). The figure~\ref{fig:paraphrase_example} depicts an example of paraphrasing an NL question template.  The SQL queries are randomly mapped to one of the four paraphrased NL questions.

\subsection{Selecting Multi-modal Answers}
\label{select_multimod_ans}
Whenever a patient is admitted to the hospital, all their treatment and medication details are immediately stored in the EHR tables (i.e. they are up-to-date). The clinical notes have elaborate details but may have outdated records, and hence less-accurate. Hence, between the two modalities, the structured records can be considered as a more reliable source of information. Therefore, in most cases the answers retrieved from structured records are considered more precise than the answers from unstructured data. This is especially true when answers directly exist in the MIMIC-III tables (i.e. non-derived relation queries). In DrugEHRQA,  questions concerning: (a) Dosage of medicine prescribed to the patient, (b) Route of medicine, (c) Form of medicine, and (d) List of medicines prescribed to the patient, are examples where answers exist directly in the MIMIC-III tables.

There are certain queries in DrugEHRQA, for which a direct answer is not available in the MIMIC-III tables (i.e. derived relation queries) because of missing data/relations. Let's consider using MIMIC III tables to answer the question: ‘What medication is the patient with an admission ID of 105104 taking for Hypoxemia?'. MIMIC-III tables contain information about the patient of interest being diagnosed with ‘Hypoxemia’. They also contain the list of medicines prescribed to the patient of interest. However, the tables may contain records (medicines) prescribed to the patient for non-Hypoxemia related conditions. In this scenario, the answer from unstructured data for such missing relations is more reliable since the answer is directly available in the clinical notes.

We have used a two-step process to generate the multi-modal answers. In the first step, an automatic method was used to retrieve the multi-modal answer. 
To automatically generate the multi-modal answers, we follow three major rules. Table~\ref{tab:rules_auto_answer} helps to explain the rules below using examples.
\begin{itemize}
\item If the answer exists in only one modality, the available answer is selected as the multi-modal answer. (1st row, Table~\ref{tab:rules_auto_answer}). 
\item Check for overlapping answers. If there is even one common answer between "Answer Structured" and "Answer Unstructured", choose the common answer. (2nd row, Table~\ref{tab:rules_auto_answer}).
\item If there are no common answers between the two modalities, choose the answer from the modality which is more reliable. (4th row, Table~\ref{tab:rules_auto_answer}). In the last row of Table~\ref{tab:rules_auto_answer}, we can observe that the answers from the two modalities are different. Since the question is a non-derived relation query, the answer from the structured database is selected as the multi-modal answer.
\end{itemize}

After generating the multi-modal answers automatically, the author manually sampled 500 queries, and cross-checked the results for the multi-modal answer. Please refer to the Appendix for further details regarding the human validation process.

\section{Analysis of the DrugEHRQA dataset}
\label{analysis_DrugEHRQA}
The SQL queries generated in the DrugEHRQA dataset can be classified into easy, medium, hard and very hard SQL queries (Refer Table~\ref{tab:complexity_levels_of_sql_queries}). The generated SQL queries were classified using the complexity determination method used in RAT-SQL~\cite{wang2020rat}. Complexities of the SQL queries are determined by factors like number of tables in the SQL query, number of conditions, presence of nesting etc. The DrugEHRQA dataset contains more complicated SQL queries (containing nested queries) than the existing text to SQL datasets  like MIMICSQL~\cite{wang2020text}

\begin{table}[ht]
    \centering
  \caption{Complexity levels of SQL queries in the DrugEHRQA dataset}
  \label{tab:complexity_levels_of_sql_queries}
  \begin{tabular}{M{3cm}| M{3cm}}
    \hline
    \thead{Difficulty levels}  & \thead{Percentage of queries} \\ 
    \hline
    Easy &  1.1\%     \\
    Medium & 39.2\%     \\
    Hard & 9.8\%  \\
    Very Hard & 49.9\%  \\

    \hline
  \end{tabular}
\end{table}

\begin{figure}[ht]
    \centering
    \includegraphics[width=0.9\linewidth]{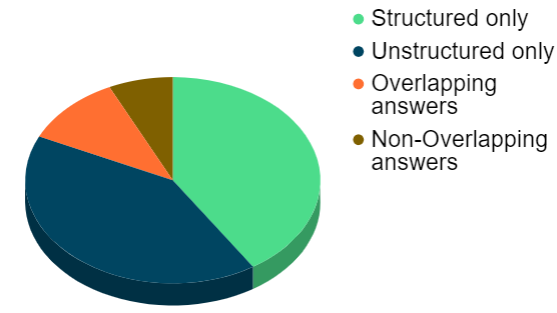}
    \caption{Distribution of DrugEHRQA dataset based on the modality or source of answers. Note that "Structured only" means that questions whose answers exist only in structured EHR data. Similarly, "unstructured only" means that questions whose answers exist only in unstructured clinical notes.}
    \label{fig:percent_answers}
\end{figure}

\begin{figure}[ht]
    \centering
   \includegraphics[scale=0.55]{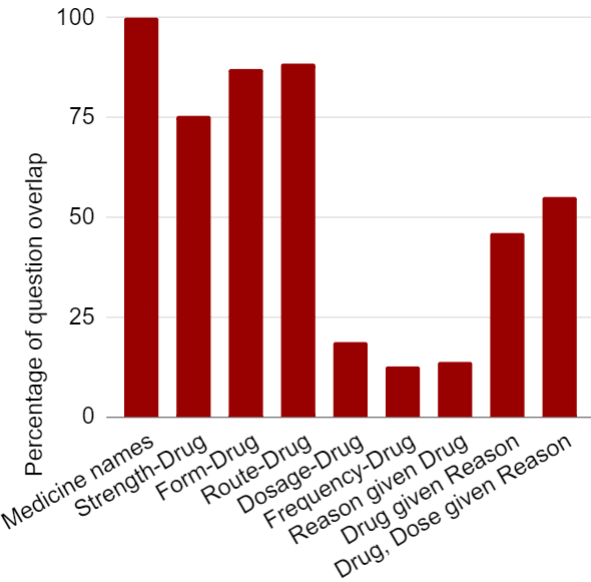}
    \caption{Percentage of questions with at least one answer-overlap from text-table QA}
    \vspace{-1em}
    \label{fig:percent_answer_overlap}
\end{figure}

DrugEHRQA contains a total of 70,381 questions along with answers from either the multi-relational tables, or the unstructured clinical data of MIMIC-III, or from both the sources. The dataset also contains an automatically generated multi-modal answer.  The DrugEHRQA dataset is diverse with respect to the source of the answers (refer to figure~\ref{fig:percent_answers}). There are questions that can be answered by individual modalities (i.e. structured or unstructured EHR). Roughly 41 \% of the drug-related queries can be answered individually by the structured data and unstructured data. It also contains questions that have answers in both the modalities (18.1\%, i.e. 12,738 queries). 11\% of the questions have at least one overlapping answer in structured and unstructured EHR data (e.g., row 2 and 3 of table~\ref{tab:rules_auto_answer}) whereas the remaining 7.1\% of the questions have distinct (or non-overlapping) answers in the two modalities (e.g., row 4 of table~\ref{tab:rules_auto_answer}). 15\% of this section of the dataset have missing relations (or information) in the structured tables, i.e. they are derived relation queries (as explained in Section~\ref{select_multimod_ans}. Hence, among the queries containing answers in both the modalities, the answers from unstructured EHR data is more reliable (than structured EHR data) for 15\% of the queries. 

 Figure~\ref{fig:percent_answer_overlap} shows the percentage of questions with at least one-answer overlap between table and text QA for the nine templates. Some of the templates like medicine names, form-drug, and route-drug have a high percentage of overlapping answers. Confidence in the accuracy of answers increases when the answers are the same, e.g.: row 2 and row 3 of Table~\ref{tab:rules_auto_answer}. Table~\ref{tab:additional_context} shows examples where multi-modal QA in EHRs can help provide additional context. We observed that the answers from the modalities were different, but the dual modalities together provide the complete answer. The answer from structured data gives the dosage in milligrams, whereas the answer retrieved from the clinical notes presents the dosage based on the number of tablets. Both of the answers are right, which can be verified from the last column, since the dosage recommended in row 1 is one 325 mg tablet, to be taken daily. In short, answers from one modality can help to provide better context to the answers retrieved from the other modality.

\section{Baseline Models}
\label{models}
We now discuss all the baseline models used for performing QA tasks on DrugEHRQA.  Among the questions containing answers in both the modalities (which comprises 18.1\% of the dataset), 60.8\% of these queries have repeated answers in the two modalities. For such questions, either modality (text QA or table QA) can be used. For questions with one or more overlapping answers, we used the non-derived relation queries for QA on structured tables, and have used unstructured data to answer the derived relation queries, as explained in Section ~\ref{select_multimod_ans}. For questions having answers in only a single modality (table or text), we use unimodal QA and separate QA baseline models to validate our QA dataset on structured EHRs and unstructured EHRs. Two existing models - TREQS~\cite{wang2020text} and RAT-SQL~\cite{wang2020rat} are used for the text-to-SQL tasks on DrugEHRQA which use MIMIC-III tables.

For questions having no common answers in the two modalities, we have developed a multimodal QA pipeline (MultimodalEHRQA). The pipeline consists of a modality selection network which predicts the modality (i.e. table or text) for a given question. If the network predicts 'text` as the modality (i.e. clinical notes), then extractive QA using BERT or ClinicalBERT is performed, predicting the  span of text from the clinical EHR notes as the answer. If the network predicts 'table` as the modality, then text-to-SQL is performed by the TREQS model. The predicted answers from the chosen modality is compared to the golden standard multimodal answer. 

\begin{figure}[ht]
    \centering
    \includegraphics[scale=0.45]{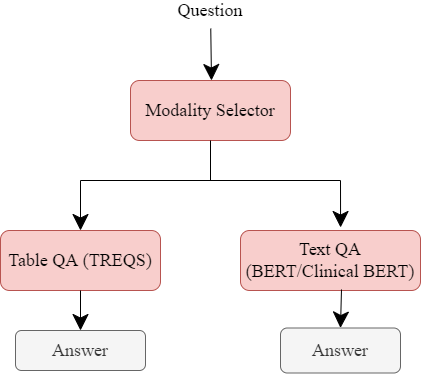}
    \caption{MultimodalEHRQA pipeline}
    \vspace{-1em}
    \label{fig:modality_selection_network}
\end{figure}

\subsection{Multimodal Selection Network}
\label{multimodal_selection_network}
The multimodal selection network uses a binary classification approach (refer Figure~\ref{fig:modality_selection_network}.
We have used BERT with a feedforward network followed by a softmax layer to predict the correct or the more reliable modality for answering the questions. 
\subsection{QA models}
\label{QA_models}
\textit{TRanslate-Edit Model for Question-to-SQL (TREQS)}~\cite{wang2020text} is a sequence-to-sequence model which generates SQL query for a given question. It makes necessary modifications to an answer with the help of an attentive copying mechanism and task-specific look-up tables. 

\textit{RAT-SQL (Relation-Aware Schema Encoding and Linking for Text-to-SQL Parsers)}~\cite{wang2020rat} was used in order to address the  more complex, nested SQL queries of the DrugEHRQA dataset. RAT-SQL uses a relation-aware self-attention mechanism to address schema encoding, schema linking, and feature representation within a text-to-SQL encoder. A self-aware attention mechanism in RAT-SQL helps to encode more complex relationships between columns and tables within the schema of the database, as well as between the question and the database schema. Because TREQS is unable to handle nested SQL queries (for 4 out of 9 templates), we adapt RAT-SQL (Relation-Aware Schema Encoding and Linking for Text-to-SQL Parsers)~\cite{wang2020rat} to test the nested templates and introduce RAT-SQL to the healthcare domain. 



\textit{BERT QA}~\cite{devlin2019bert} and \textit{ClinicalBERT QA}~\cite{alsentzer2019publicly} have gained popularity over the years for QA over unstructured data ~\cite{johnson2016mimic,soni2020evaluation}. ClinicalBERT is the clinical version of BERT pre-trained on the clinical notes of MIMIC-III. The BERT QA model is pre-trained on large datasets like BooksCorpus and English Wikipedia. The training size of Clinical BERT's corpus (roughly 50M words) is much smaller than BERT (roughly 3300M words).

\section{Experiments and Results}
\label{experiments_and_results}
Our dataset has questions with answers in a single modality (structured or unstructured EHR data), as well as questions where answers exist in both the modalities (refer to figure~\ref{fig:percent_answers}). Due to the diversity of our dataset, we have performed experiments on single-modality QA pairs, as well as for multimodal non-overlapping questions. Furthermore, we have also evaluated the overall performance of our dataset on the MultimodalEHRQA model. We define two versions of our dataset - DrugEHRQA (basic) and DrugEHRQA (extended). The difference between the two versions is that the DrugEHRQA (basic) contains all the questions of the DrugEHRQA dataset except those which have answers only in the structured EHR data and contain nested SQL queries. Note, that we can't use BERT/ClinicalBERT for these questions, as they have answers only in the MIMIC-III tables. Also, we cannot use TREQS to validate these questions, so we limit using MultimodalEHRQA to the DrugEHRQA (basic). We use RAT-SQL to determine the accuracy of the structure of the predicted SQL queries for these questions.  

\subsection{Results of Single-modal QA}
For all our experiments, divided the dataset into a 0.8/0.1/0.1 train/dev/test split.  A batch size of 16, 20, and 12 is used for TREQS, RAT-SQL, and BERT/ClinicalBERT, respectively. We trained the TREQS model for 4 epochs with a learning rate of 5e-3, grad clip of 2.0, and a maximum vocabulary size of 50,000. For the scheduler, we used a step size of 2 and step decay of 0.8 and set the minimum word frequency to 5. For RAT-SQL, we used  GloVe word embeddings~\cite{pennington2014glove}. The model was trained using GeForce RTX 2080 Ti GPUS for up to 40,000 steps with the Adam optimizer~\cite{kingma2015adam}. The same hyperparameters were used as in~\cite{wang2020rat}. For BERT and ClinicalBERT, Quadro RTX 6000 GPUs were used for training the model for 2 epochs with a learning rate of 3e-5. A doc stride of 128 was used with a max length of 384.

We use Logical Form Accuracy (Acc\_LF) and Execution Accuracy (Acc\_Ex) as evaluation metrics to test the SQL queries for the TREQS model. Logical Form Accuracy can be defined as the ratio of the number of strings matched between the ground truth and the generated SQL query, to the total number of question-SQL pairs. Execution accuracy on the other hand, represents the ratio of the number of SQL queries generated with correct answers to the total number of question-SQL pairs. Table~\ref{tab:overall_perf_treqs_ratsql} shows the performance of the TREQS model for questions having answers only in structured tables of MIMIC-III. At times, the condition value in the question may not match the table's header. The TREQS model uses a recover technique where a string matching metric, ROUGE-L, is used to search for the most similar condition value using the lookup table for every predicted SQL query. Hence, the "TREQS (with recover)" in Table~\ref{tab:overall_perf_treqs_ratsql} refers to the accuracy of the test set when the query generated using the sequence-to-sequence model is further edited to recover the exact data with the help of the table schema and look-up tables of content keywords. We observe from the table that after using recover, the overall performance improves. We also compared performance of TREQS and RAT-SQL on the non-paraphrased and paraphrased version of the dataset. Adding paraphrases, decreases LF accuracy by 0.6\% and 0.7\% respectively for TREQS and RAT-SQL. This shows that paraphrasing increases the difficulty level of questions in the dataset.

\setlength{\arrayrulewidth}{0.5mm}
\setlength{\tabcolsep}{6pt}
\renewcommand{\arraystretch}{1.1}
\begin{table}
    \centering
  \caption{Performance of DrugEHRQA on TREQS and RAT-SQL models for questions with answers only in structured EHR}
  \label{tab:overall_perf_treqs_ratsql}
  \begin{tabular}{M{4cm}|M{1cm}|M{1cm}}
    \hline
    \thead{Models} & \thead{Acc\_LF} & \thead{Acc\_EX} \\
    \hline
    TREQS (without recover) & 0.6384 & 0.6384 \\
    TREQS (with recover) & 0.648 & 0.649 \\
    RAT-SQL & 0.8723 & - \\
    \hline
  \end{tabular}
\end{table}

We also observe that the overall LF accuracy of DrugEHRQA on RAT-SQL is much higher than the TREQS model. This is because the computation of LF accuracy in RAT-SQL evaluates the predicted SQL query on all components except the condition values of the SQL query. And, prediction of the condition values in a text-to-SQL prediction task is much more challenging than predicting the other components of the SQL Query, hence the LF accuracy on RAT-SQL is higher than that of the TREQS model.

For questions having answers only in unstructured EHR data, we use exact match and F1 score as evaluation metrics. The DrugEHRQA dataset obtained an exact score of 61.363 and an F1 score of 64.307 on the test set for BERT QA (Table~\ref{tab:bert_cBert_clinical}). We obtain a marginal difference in performance between BERT and ClinicalBERT.

\setlength{\arrayrulewidth}{0.5mm}
\setlength{\tabcolsep}{6pt}
\renewcommand{\arraystretch}{1.1}
\begin{table}
    \centering
  \caption{Results of QA using BERT and Clinical BERT for questions having answers exclusively in clinical notes of MIMIC-III}
  \label{tab:bert_cBert_clinical}
  \begin{tabular}{M{1cm}|M{1cm}|M{1cm}|M{1cm}|M{1cm}}
    \hline
    \thead{ } & \thead{Dev \\Exact-\\match} & \thead{Dev F1-\\score} &\thead{Test \\Exact-\\match} & \thead{Test F1-\\score} \\
    \hline
    BERT & 60.757 & 64.968 & 61.363 & 64.307 \\
    Clinical BERT & 62.112 & 66.217 & 61.679 & 65.220 \\
    \hline
  \end{tabular}
  \vspace{-1em}
\end{table}

\subsection{Results of multimodal QA}

This section describes the performance of DrugEHRQA questions on our multimodal QA pipeline (Fig:~\ref{fig:modality_selection_network}). The modality selection network was trained on 16 epochs, with a learning rate of 1e-7 and batch size of 64.  The modality selection network gave an accuracy of 99.6\%. Following this, the questions were answered using TREQS or BERT/Clinical BERT QA, based on the predicted modality. Note, that RAT-SQL was not used to evaluate the overall exact match. This is because RAT-SQL can be used to evaluate only the structure of the SQL query template, not the exact SQL query.  

\begin{figure}[ht]
    \centering
    \includegraphics[scale=0.5]{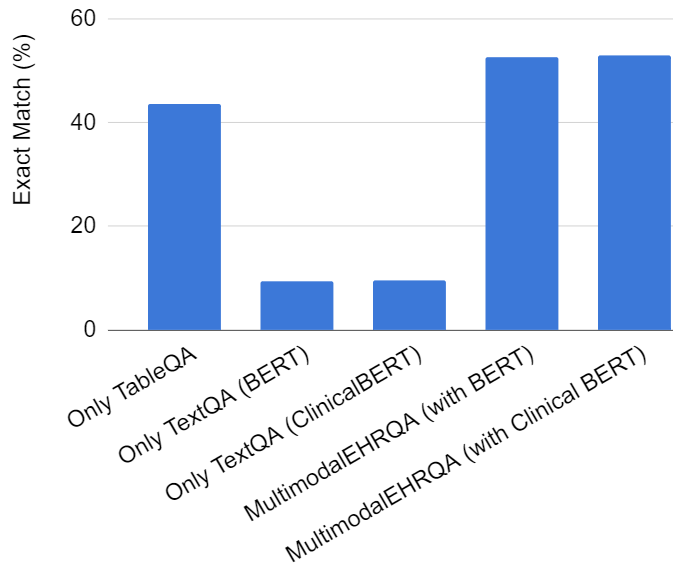}
    \caption{Exact match values of MultimodalEHRQA in comparison to single-modal QA models for questions with non-overlapping answers}
    \label{fig:Exact match values of multimodal baseline model in comparison to single-modal models for non-overlapping answers}
\end{figure}

\begin{figure}[ht]
    \centering
   \includegraphics[scale=0.5]{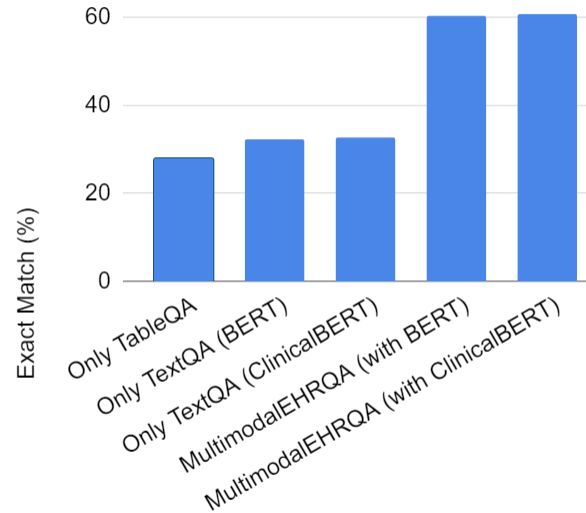}
    \caption{Overall performance of MultimodalEHRQA in comparison to single-modal QA models for the entire DrugEHRQA dataset (basic version)}
    \vspace{-1em}
    \label{fig:Exact match values of multimodal baseline model in comparison to unimodal models for all queries}
\end{figure}

Figure~\ref{fig:Exact match values of multimodal baseline model in comparison to single-modal models for non-overlapping answers} shows the performance of our MultimodalEHRQA model specifically on the questions having no overlapping answers between the two modalities. Figure~\ref{fig:Exact match values of multimodal baseline model in comparison to unimodal models for all queries} shows the exact match values of our model on the entire dataset (DrugEHRQA - basic), and compares its performance on the unimodal QA models. From both figure~\ref{fig:Exact match values of multimodal baseline model in comparison to single-modal models for non-overlapping answers} and figure~\ref{fig:Exact match values of multimodal baseline model in comparison to unimodal models for all queries}, we can see that the baseline model performs much better in comparison to the single-modal QA (Table-QA and text-QA).

\section{Reproducibility \& Limitations}
\label{limitations}
The user needs to request for credentialed access for PhysioNet\footnote{https://physionet.org/}. The user must download the MIMIC-III data, retrieve the drug relations from the ‘2018 (Track 2) Adverse Drug Event (ADE) and the Medication Extraction Challenge dataset’ from the n2c2 repository (after requesting for access to n2c2 datasets). Once this is done, the user can just replicate the steps described in the dataset generation process (Section~\ref{dataset_gen}) to produce the DrugEHRQA dataset. Even though the multimodal QA dataset generation process is automatic, without the need for long hours of annotation. But this procedure is limited only to the MIMIC-III database. The same steps cannot be reproduced for other EHR databases. In fact, MIMIC-IV~\cite{johnson2020mimic} is the latest version. But since the dataset generation process is dependent on the drug relations extracted from the challenge dataset, so our dataset generation process was limited to the MIMIC-III database. Furthermore, the diversity of the questions (or question templates) in the DrugEHRQA dataset are limited by the type of relations extracted from the challenge dataset. But despite these limitations, the DrugEHRQA dataset is an invaluable resource which will encourage further research in multimodal QA over EHRs.

\section{Broader Impact on the EHR QA research community and Future Work}
\label{broader_impact}
The DrugEHRQA dataset helps to put a spotlight on multimodal EHRs. The data in the structured and unstructured EHR may contain duplicated information (improves confidence of the answer), they may contrast each other, and may also aid in adding context to each other. This opens up new avenues of research in multimodal QA in EHRs. DrugEHRQA can be used as a benchmark model for all QA models that uses multiple EHR tables and clinical notes for information retrieval.  Since in a lot of cases, the data in structured and unstructured EHR sources helps to provide additional context to each other, another possible application of DrugEHRQA is in improving QA over structured (or unstructured), by using information or evidence from the unstructured EHR source (or structured).

\section{Conclusion}
\label{conclusion and Future Work}
To conclude, EHRs contain a large amount of up-to-date patient information in the structured databases, along with clinical notes containing elaborate details. We have introduced a novel methodology to generate a large multimodal QA dataset, containing answers from multi-relational tables and discharge summaries of a publicly available EHR database (MIMIC-III). It is the first QA dataset which contains natural language questions, SQL queries, and answers from either or both structured EHR tables and unstructured free text. Additionally, we use an automated methodology to generate the multimodal answer. Following this, human annotators verified the answers for a sampled dataset. We have also introduced a simple baseline model for multimodal QA on EHRs. In the future, we will try to work on multimodal QA models for EHRs which jointly trains the model on both table and text. Using QA over one modality to improve QA over other modalities is another promising direction of this work. The DrugEHRQA dataset introduces new horizons of research in multimodal QA over EHRs.

\section{Acknowledgements}
\label{Acknowledgements}
This work is partially support
ed by US National Science Foundation grant IIS-1526753, and Arnold and Lisa Goldberg funds.

\section{Bibliographical References}
\bibliographystyle{lrec2022-bib}
\bibliography{lrec2022-example}


\clearpage
\appendix
\section{Appendix}
\subsection{Human validation procedure}
We randomly sampled 500 queries out of the 70,380 queries of the DrugEHRQA dataset. Three human validators cross-checked the results obtained after automatically generating the selected multi-modal answers using a rule-based method described in Section \ref{select_multimod_ans} of the paper. One of the authors (a PhD student) along  with two other PhD students, working in the Computer Science Department at University of Florida verified the results.

\subsection{Annotation Guidelines
}
The same sample set of 500 was given to all the three annotators. They individually checked the results after they were explained about the annotation guidelines. The annotators were given the same general guidelines. The answers are present either in one or both the modalities (i.e. structured and unstructured data). The annotators were given the answers extracted from structured MIMIC-III tables as well as answers from unstructured clinical notes of MIMIC-III. They had to verify the labels for the selected multimodal answers manually. The list of answers in both the sources were separated by commas. The answer labels are answers from either of the two modalities. If only one of the modalities contained the answer, the available answer is selected as the multimodal answer. If answers are present in both the modalities, the annotators checked for overlapping or intersecting answers. If there was some overlap, the overlapping entries were selected as the multimodal answer. This is because if two modalities have the same answer, it increases the confidence score of that answer. If there are no answers overlapping between the two modalities, the answer retrieved from structured EHR tables was selected when the queries are of derived relation type. Examples of such derived relation type templates are: list of medicines, dosage of medicine, strength of medicine, form and route of medicine. But when the queries are of derived relation type, the answer from unstructured EHR data is selected as the “multimodal answer”. During the human verification 
process, the annotators considered some commonly occurring abbreviations, while looking for overlapping entries between the answers retrieved from the two modalities. This is listed in the table-\ref{tab:Commonly occurring abbreviations in DrugEHRQA dataset}. In case of disagreements between the human verification results, the annotators discussed the conflicting answers and again rechecked those samples together and came to a common conclusion based on majority voting.
\setcounter{table}{0}
\renewcommand{\thetable}{A\arabic{table}}

\setlength{\arrayrulewidth}{0.5mm}
\setlength{\tabcolsep}{6pt}
\renewcommand{\arraystretch}{1.1}
\begin{table}
    \centering
  \caption{Commonly occurring abbreviations in DrugEHRQA dataset}
  \label{tab:Commonly occurring abbreviations in DrugEHRQA dataset}
  \begin{tabular}{M{2cm}|M{4cm}}
    \hline
    \thead{Abbreviation} & \thead{Terms} \\
    \hline
    TAB & TABLET \\
    CAP & CAPSULE \\
    IH & INHALATION \\
    SYR & SYRINGE \\
    IV & INTRAVENOUS \\
    PO/NG & PO \\
    \hline
  \end{tabular}
\end{table}

Only 0.2\% of the queries were incorrect. This proved that our rule-method is extremely efficient.

\subsection{Dataset Accessibility}
Since the dataset contains patient sensitive information and due to license issues of the database/derived datasets, the DrugEHRQA dataset is submitted at two places. The datasets will be hosted by Physionet and n2c2 respository. The DrugEHRQA dataset will be available to the public through credentialed access to Physionet and n2c2. The natural language question, SQL queries, and answers from structured EHR data is submitted in Physionet, whereas the NL questions and answers from unstructured EHR data is submitted in the n2c2 repository. The dataset submitted to Physionet can now be accessed from https://physionet.org/~\cite{bardhandrugehrqa}.\footnote{The question-answers from unstructured EHR data, submitted to n2c2 is still under review and will be available shortly in 
https://www.i2b2.org/NLP/DataSets/}. 

\subsection{Question templates with examples}
This section uses Table~\ref{tab:NL_question_template},  table~\ref{tab:rules_auto_answer}, table~\ref{tab:additional_context}, and table~\ref{tab:app_tab1} to list the different question templates, followed by some examples. The Table~\ref{tab:NL_question_template} describes the different question templates of the dataset derived from the drug attributes and entities in the "2018 (Track 2) Adverse Drug Event (ADE) and the
Medication Extraction Challenge dataset". Table~\ref{tab:rules_auto_answer} and Table~\ref{tab:additional_context} displays examples from the dataset where the two modalities (i.e. structured and unstructured EHR data) contain similar answers (for example, 2nd and 3rd row of table~\ref{tab:rules_auto_answer}), when the two modalities contain conflicting or dissimilar answers (example: 4th row of table~\ref{tab:rules_auto_answer}), and also shows examples where the answers retrieved from structured and unstructured EHR data complement each other (for example, row 1 and 2 of table~\ref{tab:additional_context}). The rules described in Section~\ref{select_multimod_ans} was used to obtain the multimodal answers. Finally, the table~\ref{tab:app_tab1} lists the NL question templates, its corresponding SQL query templates, and their difficulty level. 

\setlength{\arrayrulewidth}{0.5mm}
\setlength{\tabcolsep}{6pt}
\renewcommand{\arraystretch}{1.5}
\begin{table*}[ht]
    \centering
  \caption{NL Question templates derived from drug-related entities and attributes extracted from the clinical notes using the n2c2 dataset, along with examples}
  \label{tab:NL_question_template}
  \begin{tabular}{M{2.5cm}|M{3.5cm}|M{6cm}}
    \hline
    \thead{Drug attributes\\ and entities} & \thead{Examples} & \thead{NL Question templates} \\
    \hline
    Drug & Lithium Carbonate, Propafenone & What are the list of medicines prescribed to the patient\\
    Strength-Drug & (300mg, Lithium Carbonate) & What is the drug strength of |drug| \\
    Form-Drug & (Tablet, Propafenone) & What is the form of |drug|  \\
    Route-Drug & (PO, Metoprolol Tartrate) & What is the route of administration for the drug |drug|  \\
    Dosage-Drug & (One tablet, Bactrim) & What is the dosage of |drug| prescribed to the patient  \\
    Frequency-Drug & (14 day, Zosyn) & How long has the patient been taking |drug| \\
    Reason-Drug & (Constipation, Polyethylene Glycol) & Why is the patient been given |drug|  \\
    Reason-Drug & (Polyethylene Glycol, Constipation) & What is the medication prescribed to the patient for |problem|\\
    Reason-Drug, Dosage-Drug & (Constipation, Polyethylene Glycol), (300mg , Polyethylene Glycol) & List all the medicines and their dosages prescribed to the patient for |problem| \\

    \hline
  \end{tabular}
\end{table*}

\setlength{\arrayrulewidth}{0.5mm}
\setlength{\tabcolsep}{6pt}
\renewcommand{\arraystretch}{1.5}
\begin{table*}[ht]
    \centering
  \caption{Information in structured and unstructured EHR providing additional context to each other. Note that the field ‘Answer Unstructured’ is the direct answer extracted from unstructured data with the help of the n2c2 dataset, and the field ‘Phrases from clinical notes’ are the lines of text in the discharge summary from which the answer is extracted.}
  \label{tab:additional_context}
  \begin{tabular}{M{3.5cm}|M{3cm}|M{2cm}|M{3.5cm}}
    \hline
    \thead{NL Questions} & \thead{Answer from \\structured} & \thead{Answer from\\ unstructured} & \thead{Phrases from \\ clinical notes} \\
    \hline
    WHAT IS THE DOSE OF ASPIRIN THAT THE PATIENT WITH ADMISSION ID = 142444 HAS BEEN PRESCRIBED & 325MG,300MG & ONE (1) & 325 mg Tablet Sig: One (1) Tablet PO DAILY (Daily). 5. Acetaminophen 325 mg Tablet Sig: One (1) Tablet PO Q6H (every 6 hours) as needed. \\
    
    LIST ALL THE MEDICINES AND THEIR DOSAGES PRESCRIBED TO THE PATIENT WITH ADMISSION ID = 105014 FOR POLYMYALGIA RHEUMATICA & PREDNISONE: 20 MG, TACROLIMUS: 4 MG, MYCOPHENOLATE MOFETIL: 1000 MG, TACROLIMUS: 4 MG, TACROLIMUS: 5 MG, MYCOPHENOLATE MOFETIL: 500 MG &  PREDNISONE: ONE (1) & 20 mg Tablet Sig: One (1) Tablet PO DAILY (Daily). \\
    \hline
  \end{tabular}
\end{table*}

\setlength{\arrayrulewidth}{0.5mm}
\setlength{\tabcolsep}{6pt}
\renewcommand{\arraystretch}{1.5}
\begin{table*}[ht]
    \centering
  \caption{Rules for automatic multi-modal answer retrieval}
  \label{tab:rules_auto_answer}
  \begin{tabular}{M{6.75cm}|M{2cm}|M{2cm}|M{1.5cm}}
    \hline
    \thead{Question} & \thead{Answer from\\\ Structured} & \thead{Answerfrom\\\ Unstructured} & \thead{Multi-modal \\ answer} \\
    \hline
    WHAT IS THE MEDICATION PRESCRIBED TO THE PATIENT WITH ADMISSION ID 111160 FOR PAIN & -- & MORPHINE & MORPHINE \\
    WHAT IS THE DRUG STRENGTH OF SIMETHICONE PRESCRIBED TO THE PATIENT WITH ADMISSION ID 125206 & 80MG TABLET & 80 MG & 80MG TABLET \\
    HOW LONG HAS THE PATIENT WITH ADMISSION ID = 187782 BEEN TAKING VANCOMYCIN & 14 DAYS & 14 DAYS & 14 DAYS \\
    WHAT IS THE DRUG STRENGTH OF FUROSEMIDE PRESCRIBED TO THE PATIENT WITH ADMISSION ID 100509 & 40MG/4ML VIAL & 10 MG & 40MG/4ML VIAL \\
    \hline
  \end{tabular}
\end{table*}

\clearpage
\onecolumn 
\setlength{\arrayrulewidth}{0.5mm}
\setlength{\tabcolsep}{6pt}
\renewcommand{\arraystretch}{1.5}
\begin{longtable}{M{0.50cm}|M{3cm}|M{7cm}|M{1.25cm}}
    \caption{Templates and their level of difficulty}
        \label{tab:app_tab1}\\
    \hline
        \thead{Sl.\\No} &  \thead{NL Question\\Template}  & \thead{SQL Query Template} & \thead{Difficulty\\Level} \\ 
        \hline
        1. & What are the list of medicines prescribed to the patient with admission id |hadm\_id| & SELECT PRESCRIPTIONS.DRUG FROM PRESCRIPTIONS WHERE PRESCRIPTIONS.HADM\_ID = |hadm\_id| & Easy \\ 
        2. & What is the drug strength of |drug| prescribed to patient  with admission id |hadm\_id| & SELECT PRESCRIPTIONS.PROD\_STRENGTH FROM PRESCRIPTIONS WHERE PRESCRIPTIONS.HADM\_ID = |hadm\_id| AND PRESCRIPTIONS.DRUG = |drug| & Medium \\
        3. & What is the form of |drug| prescribed to patient with admission id |hadm\_id| & SELECT PRESCRIPTIONS.FORM\_UNIT\_DISP FROM PRESCRIPTIONS WHERE PRESCRIPTIONS.DRUG = |drug| AND PRESCRIPTIONS.HADM\_ID = |hadm\_id| & Medium \\ 
        4. & What is the route of administration for the drug |drug| for patients with admission id = |hadm\_id| & SELECT PRESCRIPTIONS.ROUTE FROM PRESCRIPTIONS WHERE PRESCRIPTIONS.DRUG = |drug| AND PRESCRIPTIONS.HADM\_ID = |hadm\_id| &  Medium \\
        5. & What is the dosage of |drug| prescribed to the patient with admission id = |hadm\_id| & SELECT PRESCRIPTIONS.DOSE\_VAL\_RX, PRESCRIPTIONS.DOSE\_UNIT\_RX FROM PRESCRIPTIONS WHERE PRESCRIPTIONS.HADM\_ID = |hadm\_id| AND PRESCRIPTIONS.DRUG = |drug| & Medium \\
        6. & How long has the patient with admission id = |hadm\_id| been taking |drug| & SELECT SUM(PRESCRIPTIONS.DURATION\_IN\_DAYS) FROM PRESCRIPTIONS WHERE PRESCRIPTIONS.HADM\_ID = |hadm\_id| AND PRESCRIPTIONS.DRUG = |drug| GROUP BY PRESCRIPTIONS.HADM\_ID, PRESCRIPTIONS.DRUG & Hard \\
        7. & Why is the patient with admission id = |hadm\_id| been given |drug| & SELECT L3.SHORT\_TITLE FROM D\_ICD\_DIAGNOSES AS L3 WHERE L3.ICD9\_CODE IN  (SELECT L1.ICD9\_CODE  FROM DIAGNOSES\_ICD AS L1 INNER JOIN PRESCRIPTIONS AS L2 ON L1.HADM\_ID = L2.HADM\_ID WHERE L1.HADM\_ID = |hadm\_id| AND L2.DRUG = |drug|) & Very hard \\
        8. & What is the medication prescribed to the patient with admission id = |hadm\_id| for |problem| & SELECT Y.DRUG FROM PRESCRIPTIONS AS Y WHERE Y.HADM\_ID = (SELECT L1.HADM\_ID FROM DIAGNOSES\_ICD AS L1 INNER JOIN D\_ICD\_DIAGNOSES AS L2 ON L1.ICD9\_CODE = L2.ICD9\_CODE WHERE L1.HADM\_ID = |hadm\_id| AND L2.LONG\_TITLE =  |problem|) & Very hard \\
        9. & List all the medicines and their dosages prescribed to the patient with admission id = |hadm\_id| for |problem| & SELECT Y.DRUG, Y.DOSE\_VAL\_RX, Y.DOSE\_UNIT\_RX FROM PRESCRIPTIONS AS Y WHERE Y.HADM\_ID = (SELECT L1.HADM\_ID FROM DIAGNOSES\_ICD AS L1 INNER JOIN D\_ICD\_DIAGNOSES AS L2 ON L1.ICD9\_CODE = L2.ICD9\_CODE WHERE L1.HADM\_ID = |hadm\_id| AND L2.LONG\_TITLE = |problem|) & Very hard \\

        \hline
    
\end{longtable}

\end{document}